# Breast density in MRI: an AI-based quantification and relationship to assessment in mammography

Yaqian Chen[1,6], Lin Li[1,6], Hanxue Gu[1], Haoyu Dong[1], Derek L. Nguyen[2],

Allan D. Kirk[3], Maciej A. Mazurowski[1,2,4,5], E. Shelley Hwang[3*]

1 Department of Electrical and Computer Engineering, Duke University
2 Department of Radiology Duke University School of Medicine
3 Department of Surgery Duke University School of Medicine
4 Department of Biostatistics & Bioinformatics, Duke University
5 Department of Computer Science, Duke University
6 These authors contributed equally to this work as first authors.
*This is the corresponding author, shelley.hwang@duke.edu

**Abstract**

Mammographic breast density is a well-established risk factor for breast cancer. Recently there has been interest in breast MRI as an adjunct to mammography, as this modality provides an orthogonal and highly quantitative assessment of breast tissue. However, its 3D nature poses analytic challenges related to delineating and aggregating complex structures across slices. Here, we applied an in-house machine-learning algorithm to assess breast density on normal breasts in three MRI datasets. Breast density was consistent across different datasets (0.104 - 0.114). Analysis across different age groups also demonstrated strong consistency across datasets and confirmed a trend of decreasing density with age as reported in previous studies. MR breast density was correlated with mammographic breast density, although some notable differences suggest that certain breast density components are captured only on MRI. Future work will determine how to integrate MR breast density with current tools to improve future breast cancer risk prediction.

**Keywords:** breast density, breast MRI, breast imaging

## 1. Introduction

As a risk factor for breast cancer and a key evaluable factor in breast imaging [1, 2, 3, 4], mammographic breast density (MBD) has received increasing attention from clinicians, researchers, patients, and policy makers [1, 2, 5, 6]. Numerous studies have provided quantitative and qualitative analyses of breast density, using different imaging modalities including mammography, ultrasound, and MRI, to assess the utility of breast density for cancer detection and prediction of patient outcomes [4, 7, 8, 9, 10].

Both mammography and MRI are primary imaging modalities used for breast cancer screening and diagnosis. Mammography, while being widely accessible and cost-effective, is known to have limited sensitivity and specificity in patients with dense breast tissue [11]. For instance, the similarity in appearance between breast cancer and dense breast tissue can lead to radiologists mistaking subtle cancer for normal breast tissue. This limitation also makes high-precision quantitative analysis challenging, leading to four subjectively determined categories of breast density reported in current clinical practice, from "Almost entirely fatty" to "Extremely dense," which are subject to high inter-reader variability [12, 13]. The misclassification of breast density may have clinical implications, such as inaccurate lifetime breast cancer risk calculations, and may reduce the performance of MBD in risk prediction [14].

Conversely, MRI offers advantages such as enhanced sensitivity for breast cancer detection [11] and high contrast resolution [15]. However, the utilization of MRI in assessing breast density remains under-explored compared to mammography [5, 9, 10] largely due to the cost and limited availability of data compared to mammography, and the high burden of manually annotating breast dense tissue masks [16]. Common segmentation methods, like C-fuzzy clustering, lack precision and often require manual correction [17, 18]. While some deep learning-based MRI breast dense tissue segmentation models exist, they face challenges related to limited public accessibility, poor reproducibility, and insufficient validation across diverse institutional datasets [10].

To address this gap, we developed a deep-learning-based segmentation model, building on prior research [16], specifically designed for dense tissue segmentation in breast MRI. The model achieved a Dice similarity coefficient (DSC) of 0.88 for fibroglandular tissue (FGT) and 0.95 for breast, which are separately compared to radiologist-validated annotations in real-world clinical settings. Here, we test the generalizability of our automatic MRI breast density calculation pipeline by assessing the method's adaptability and effectiveness across three different datasets which varied by exam year, population, and scanner features. Furthermore, understanding the distribution of breast density across different demographic groups, such as age, remains a key purpose in our study. Previous studies have highlighted the importance of these variations, yet the detailed comparison between MR and mammogram-based breast density measurements has not been fully explored. Therefore, in this study we will also investigate the correlation between MRI and mammogram imaging modalities in measuring breast density.

We will release our code upon paper acceptance.

## 2. Result

In this study, we analyzed 1999 patients among three distinct breast MRI datasets. Table 1 provides a demographic overview of patients from the analyzed datasets: ISPY2, DBC-MRI dataset, and Internal dataset at our institution.

| | | | **ISPY2** | **DBC-MRI** | **Internal Dataset** |
|---|---|---|---|---|---|
| | Number of patients | | 717 | 890 | 392 |
| | Number of exams | | 717 | 890 | 392 |
| Demography | | Age | 48.8 (24 - 73) | 52.8 (21 - 69) | 49.8 (28 - 77) |
| | Ethnicity | Hispanic or Latino | 13.33% | 1.91% | 5.62% |
| | | Not Hispanic or Latino | 86.61% | 98.09% | 94.38% |
| | Race | White | 79.22% | 70.22% | 79.59% |
| | | African American | 12.53% | 22.58% | 12.50% |
| | | Asian | 6.42% | 4.27% | 3.06% |
| | | American Indian | 0.45% | 0.45% | 1.02% |
| | | Other | 0.97% | 5.31% | 3.83% |

Table 1: Patient Age and Race-ethnicity Characteristics for ISPY2, DBC-MRI, and Internal Dataset

The table lists the number of patients and exams; average age and age range; patients' ethnicity, and the patients' racial composition. One of the datasets (Dataset 3) consisted of internal institutional data which included corresponding mammogram images which were used to compare MR breast density calculations with radiologists' evaluations of MBD.

Our segmentation method outperforms the other architectures by achieving the highest DSC (88.79%) and the lowest HD (31.28), as shown in Table 2. Our MR-based breast density analysis reveals a similar density distribution across the three datasets and confirms that breast density generally decreases with age. Furthermore, in a subset of patients, MRI-derived values show a notable correlation with mammographic density. These findings underscore the reliability of our method in capturing key breast density characteristics across diverse populations.

| Architecture | DSC (%) ↑ | HD (pixels) ↓ |
|---|---|---|
| U-Net3D | 81.65 | 76.69 |
| nnU-Net | 82.30 | 50.16 |
| 3D V-Net | 88.79 | 31.28 |

Table 2: Comparison of segmentation performance across different 3D architectures. Metrics include Dice Similarity Coefficient (DSC ↑) and Hausdorff Distance (HD ↓). The best performance among the architectures is highlighted in bold.

2.1 Breast density distribution for three datasets

Figure 1 demonstrates the histogram of breast density values of the normal breast from the three different datasets. The histogram indicates a similar MR density distribution among all three datasets, with most breast density values falling between 0.0 and 0.2, with a rapid decline in frequency for ratios above 0.2, and very few past 0.4. We observed a consistent breast density distribution across different datasets, with the average and std. dev. of $0.104 \pm 0.092$ for ISPY2 dataset $0.114 \pm 0.096$ for DBC-MRI dataset, and $0.114 \pm 0.112$ for the internal dataset. The average MR breast density volume across all datasets was $0.111 \pm 0.098$.

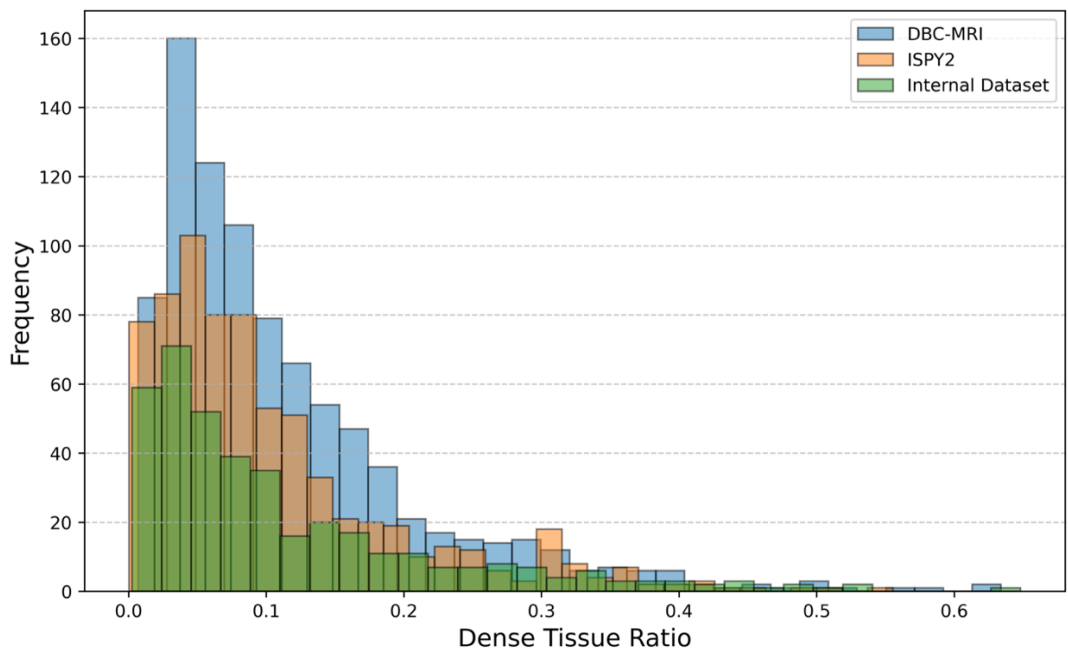

Figure 1: Breast Density Distribution for Three Different Datasets: ISPY2, DBC-MRI, and Internal Dataset.

## 2.2 Breast density distribution across age categories

Box plots in Figure 2 illustrate the distribution of breast density across various age categories in each dataset.

As shown in Figure 2, breast density generally decreases with increasing age across all datasets, with on average 0.21±0.14 for the 20-29 age group and 0.06±0.05 for the 70-79 age group. Younger age groups (20-29 and 30-39) show greater breast density variability with standard deviation of 0.14 and 0.11 respectively. In contrast, the older age groups (60-69 and 70-79) show much lower variability, with standard deviations of 0.06 and 0.05, respectively. The box plot for the 80-89 age group only contains data from the DBC_MRI dataset, showing a slightly higher average breast density (0.09±0.07) compared to the 70-79 age group (0.06±0.05). This trend aligns with previous findings [19, 20, 21, 22]. Specifically, Ohmaru, A. et al., Mustafa, W. et al., and Jiang, S. et al. demonstrated the inverse relationship between breast density and age through mammography. Mustafa, W. et al. also found the same conclusion using ultrasound, while Perera, D. et al. confirmed the consistency of these findings using optical breast spectroscopy (OBS), dual-energy X-ray absorptiometry (DXA), and mammography.

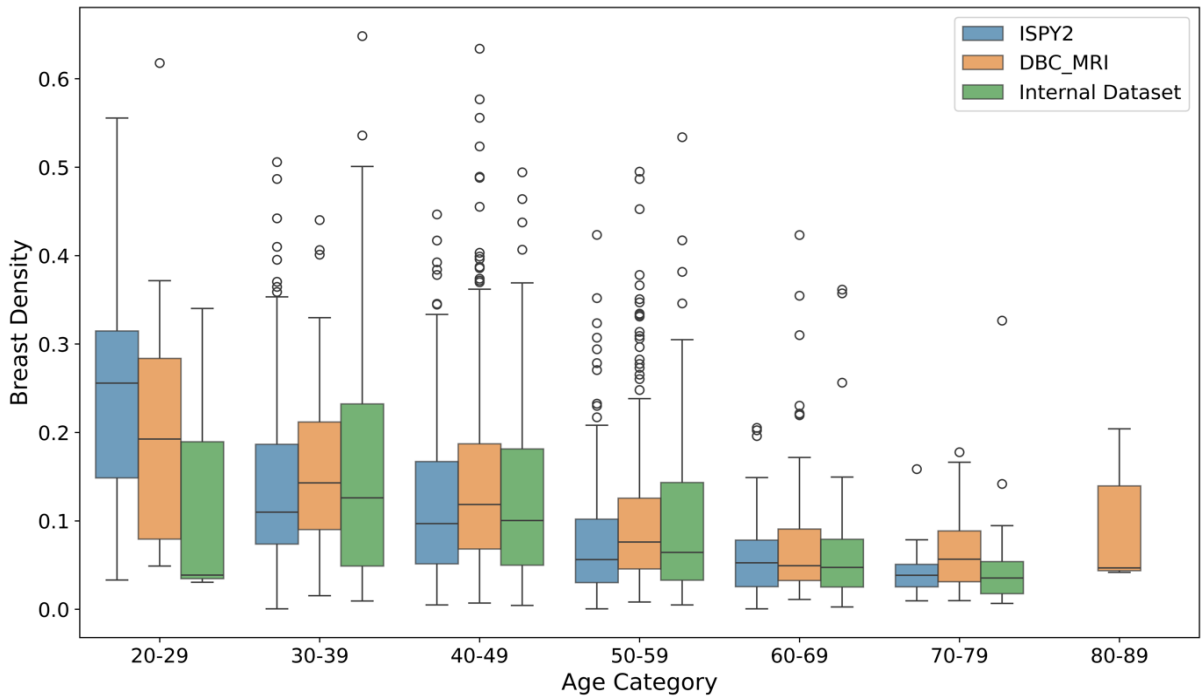

Figure 2: Breast Density Distribution According to Age Categories on ISPY2 Dataset, DBC_MRI, and Internal Dataset. The x-axis shows different age categories, ranging from 20-29 to 70-79 years. The y-axis represents the breast density, showing the proportion of dense tissue in the breast.

## 2.3 Relationship between breast density assessment in MRI and mammography

To compare the breast density given by radiologist-reviewed mammogram report and that from our auto-segmentation model, Figure 3 illustrates the breast density distribution for different density categories provided in mammography reports. The x-axis shows different density categories extracted from the mammogram report, with four levels: "Extremely dense", "Heterogeneously dense", "Scattered areas of fibroglandular density", and "Almost entirely fatty". The y-axis shows the MRI breast density extracted using our deep learning-based method. The analysis is based on MRI and mammogram from the internal breast dataset where this data was available.

We found a clear trend of decreasing breast density from the "Extremely dense" to the "Almost entirely fatty" categories, with an average breast density of 0.27 ± 0.13 for the "Extremely dense" group and 0.019 ± 0.02 for the "Almost entirely fatty" group (Figure 3). We also employed two correlation metrics to assess the relationship and agreement between MRI- and mammography-derived densities. The Spearman's Rank Correlation Coefficient was calculated to be 0.750, with a highly significant p-value of $6.64 \times 10^{-72}$, indicating a strong positive monotonic relationship between the two variables. Similarly, the Kendall's Tau Correlation Coefficient was found to be 0.622, with a p-value of $6.97 \times 10^{-58}$, further supporting a strong and statistically significant association. These results confirm a high level of agreement, suggesting that as the MRI-derived density categories increase, the mammography-derived densities also show a consistent increasing trend.

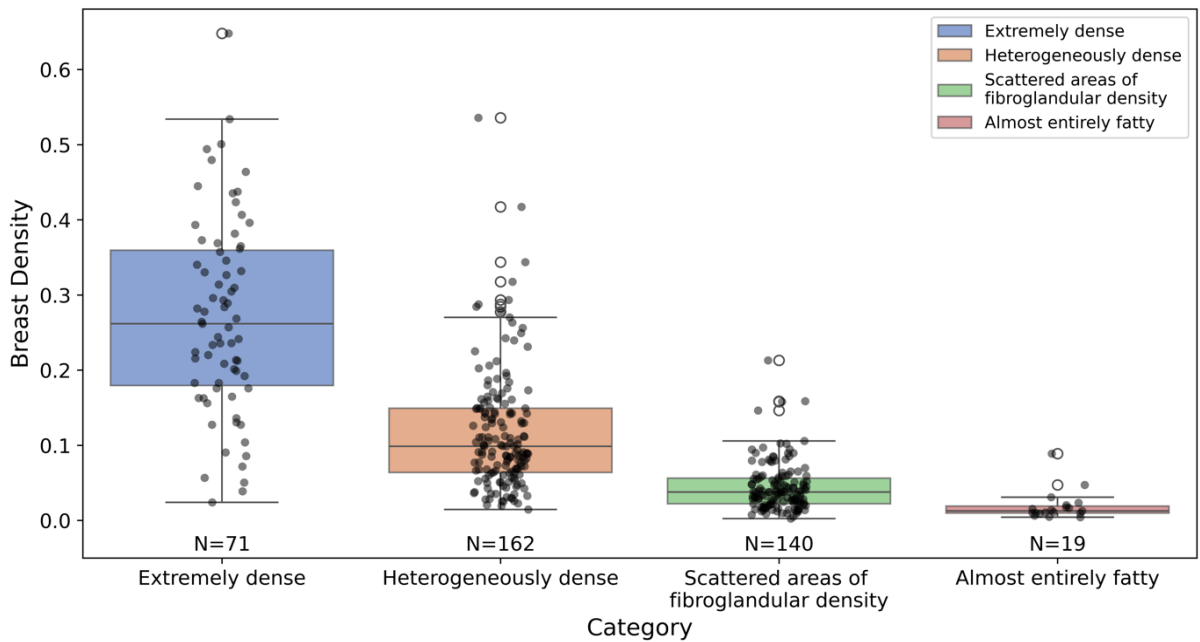

Figure 3: Breast Density Distribution for Different Density Level from Mammogram Report on our Internal Breast Dataset

## 3 Discussion

In this study, we employed an automated algorithm for breast density assessment MRI using a deep learning-based approach for breast volume and dense tissue segmentation. This automated breast density calculation was applied to three different MRI datasets in both high-risk screening and breast cancer work-up settings. We compared the breast density distribution across these MRI datasets and analyzed the breast density measurements in relation to patient demographics. Importantly, we also compared the breast density measurements obtained from MRI with the gold standard breast density evaluation from each patient's prior mammogram.

An important observation is that despite the differences between the three datasets with respect to patient population, imaging protocol, and MR equipment, the overall distributions of MR density based on our algorithm did not show notable differences. It is important to note that one of the three datasets had the high-risk screening indication meaning that the patients did not have a known cancer at the time that the study was obtained. This similarity persisted within specific age groups except for the 20-29 group where the differences were somewhat larger. Even though these three datasets were collected at different institutions for different clinical indications, the breast density distributions remained consistent.

Previous mammographic screening research has shown that breast density decreases significantly with age, influencing breast cancer risk [23]. In our analysis of three datasets measuring MR breast density, we confirmed this finding, with 92.69% difference in mean MR breast density between the highest MBD and lowest MBD groups. Although, box plot for the 80-89 age group only contains data from the DBC_MRI dataset, showing a slightly higher average breast density ($0.09\pm0.07$) compared to the 70-79 age group ($0.06\pm0.05$), this inconsistency is likely due to the limited number of cases in the 80-89 age group, which includes only six exams. Notably, we identified a more pronounced decrease in breast density between the 40-49 and 50-59 age groups compared to other age ranges. This significant drop was consistently observed across all three datasets and likely reflects a change in breast density associated with menopause [24].

When comparing the mammographic breast density categories extracted from mammography radiology report based on radiologists' interpretation of digital breast tomosynthesis (DBT) mammograms with the automatically measured breast density in MRI, a consistent correlation was observed. As expected, cases identified as having higher MBD also exhibited higher breast density ratios in MRI, as determined by our quantitative calculations. However, the assessments from MRI and mammography are not perfectly aligned, with two key observations noted. Firstly, we observed some cases of notable discrepancy between mammographic and MRI breast density specifically for "heterogeneously dense" and "extremely dense". Secondly, discrepancy between MRI and mammogram is more pronounced in "heterogeneously dense" and "extremely dense" categories than in the "Almost entirely fatty" or "Scattered areas of fibroglandular density" categories. For the first phenomenon, we believe that an important reason for this discrepancy is the previously documented [25, 26], notable inconsistency in radiologists' reporting of breast density categorizations in mammograms. The second reason are the differences between the tomosynthesis mammograms and MRI quantifications as while tomosynthesis mammography is pseudo three dimensional, MRI is a truly three dimensional which allows for different manner of quantification.

For instance, most of the breast MRI volumes categorized as "Extremely dense" had a measured MRI breast density below 0.5, which differs from the density assessments made by radiologists on DBT, where this MRD category consists of patients whose breast tissue is diffusely dense throughout. Although fibroglandular tissue as shown in Figure 4 (4(a)(ii)-4(a)(iv), and 4(b)(ii)-4(b)(iv)) appears to visually occupy a significant portion of the breast, it comprises a smaller proportion relative to fat than what the radiologist's eyes might perceive. Additionally, as the density is measured in three dimensions, there are regions of the breast, represented by slices far from the central slice which depict parts of the breast almost entirely consisting of fat. The degree of this phenomenon varies as demonstrated by the difference between the case in Fig 5a and 5b. Specifically, panels 4(a)(i) and 4(b)(i) illustrate how the density changes from slice to slice for two cases demonstrating a significantly larger portion of fatty slices for one of the cases despite a similar density in the central slice. This results in sometimes lower than expected and varying measurements of density, which will allow future discrimination of MR density subgroups within the single "Extremely dense" MRD category, which may have different implications for breast cancer risk.

For the second phenomenon, the discrepancies between mammography and MRI in assessing these categories highlight the tendency of mammography-based assessment to overestimate breast density compared to MRI-based assessment. This overestimation is likely due to the projectional nature of mammography, where overlapping fibroglandular tissue can give the appearance of higher density than is present.

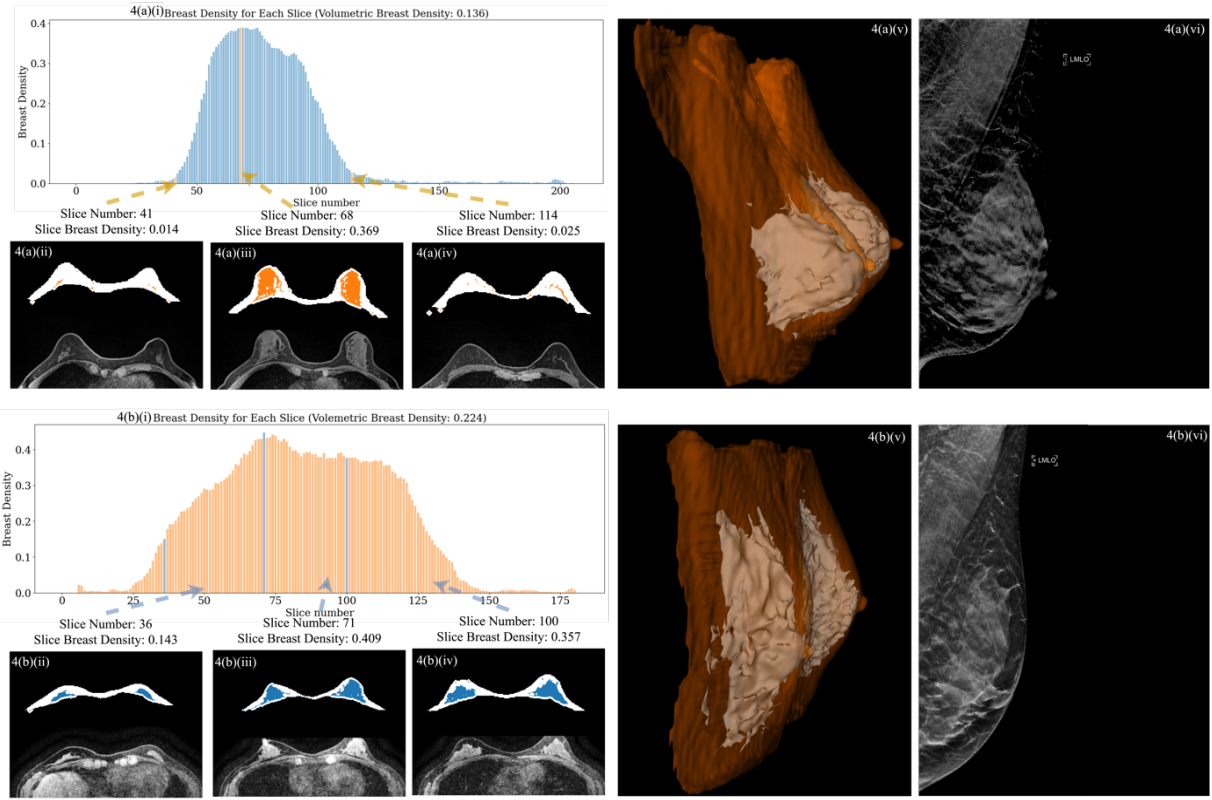

Figure 4: Mammogram and MRI slice analysis for two patients, showing breast density distribution and 3D breast tissue rendering. Panels 4(a)(i)–4(a)(vi) represent Patient A, and Panels 4(b)(i)–4(b)(vi) represent Patient B. Panels 4(a)(i) and 4(b)(i) display MR density distributions, with volumetric densities of 0.136 for Patient A and 0.224 for Patient B. Panels 4(a)(ii)–4(a)(iv) and 4(b)(ii)–4(b)(iv) show dense tissue and breast masks at selected slices: an edge near 20%, the maximum density slice, and another edge near 55%. These slices were based on Patient A's distribution and matched for Patient B for comparison. Panels 4(a)(v) and 4(b)(v) provide 3D renderings with dense tissue in grey and the full breast volume in transparent orange, while Panels 4(a)(vi) and 4(b)(vi) present mammograms labeled "Extremely dense."

In contrast, underestimation is less likely. Even if dense tissue is well-aligned across slices, a true 3D modality like MRI would not depict the tissue as less dense than it is. So, when DBT shows sparse fibroglandular tissue, it's usually a reliable indication of true low density, which MRI would confirm.

Therefore, based on this observation, we believe that MRI, with its true 3D imaging capability, provides a more accurate assessment of the actual volume of dense tissue. MRI-based breast density evaluation can help identify distinct subgroups within the "Extremely dense" category, which may be associated with varying levels of cancer risk. In contrast, for the "Almost entirely fatty" category, both mammography and MRI generally show consistent results, with MRI's precise volumetric measurements confirming the low-density classification.

Also, compared with existing mammography-based approaches, our MRI-based breast density assessment shows promising advantages in both segmentation accuracy and BI-RADS classification performance. While mammogram-based models have reported Dice Coefficients (DCS) of 88%, 77%, 76%, and 63% for dense tissue segmentation [27, 28, 29, 30], our model achieves a DCS of 88%. For overall breast segmentation, our model outperforms the 92.5% DCS reported in [31], achieving 95%. Furthermore, even with a simple SVM classifier using only the MRI-derived breast density from our model, we achieved a four-class BI-RADS classification accuracy of 73.42% on the test set, based on an 80/20 train-test split and using radiologist assessments as ground truth. This performance is comparable to the reported average accuracy of 71.3% (71.7% for MLO and 71.0% for CC) in [32]. In the binary classification setting—distinguishing between "scattered density" and "heterogeneously dense" categories—the SVM model achieved an AUC of 91%, comparable to the reported values of 94% and 91.5% (95% for MLO and 88% for CC) in [33, 34]. These results underscore the potential of MRI in improving diagnostic performance due to its high sensitivity [35].

The model also has potential limitations, particularly in accounting for the spatial distribution of dense tissue. Although 3D visualizations (Figure 4) illustrate tissue shape, a quantitative measure of distribution is lacking. This is an important consideration, as uneven tissue distribution may influence risk assessment outcomes. For instance, localized clusters of dense tissue might carry different risk implications compared to evenly distributed dense tissue, even if the overall volumetric ratio is similar. Another limitation is the potential barrier to widespread adoption due to the accessibility and cost-effectiveness of MRI. However, breast MRI is increasingly utilized in specific high-risk populations, such as individuals with BRCA mutations or those with extremely dense breast tissue [36, 37]. In addition, increasing implementation of "fast MRI" protocols have made breast MRI more affordable for both patients and health systems. The superior sensitivity of MRI, particularly in detecting abnormalities in dense breasts, underscores its potential utility in enhancing breast density-based risk prediction models. Another limitation of the study is that the patient population undergoing breast MRI are typically high-risk screening patients or patients undergoing extent of disease evaluation and not the general screening population. Therefore, this may limit the generalizability of our model's results to only the subset of patient populations undergoing breast MRI.

In summary, our analysis demonstrates a high level of consistency in breast density distribution as measured by MRI across all three datasets, affirming the reliability of our model in when applied to the subset of patient population undergoing breast MRI and MR imaging platforms. As expected, we observed a significant reduction in MR breast density with increasing age across all datasets which is consistent with the previous studies [38]. Finally, we also observed and analyzed discrepancies between mammographic and MRI assessments of breast density which provide opportunities to explore settings in which MR may provide additional, or orthogonal data to that assessed by mammography. The MR-based breast density algorithm presented here has the potential to combine MR data in a highly quantitative and reproducible method across diverse datasets, thus greatly facilitating future investigations to increase the utility of MR breast density for breast cancer risk assessment.

# 4 Methods

## 4.1 Datasets

In this section, we introduce the three datasets utilized in this study. The datasets vary in size, imaging protocols, and patient demographics, contributing to a comprehensive evaluation of breast tissue characteristics. Additional characteristics of each dataset are provided below. The research protocol was approved by the Duke Health System Institutional Review Board (IRB), we obtained the waiver for the informed consent due to the retrospective nature of the research.

**Dataset 1: I-SPY 2 Breast Dataset:** The I-SPY 2 (Investigation of Serial Studies to Predict Your Therapeutic Response with Imaging and Molecular Analysis 2) [39, 40] breast dataset includes DCE-MRI data for 719 patients with early-stage breast cancer who were adaptively randomized to drug treatment arms between 2010 and 2016. These images, acquired at over 22 clinical centers, focus on patients undergoing neoadjuvant chemotherapy (NAC). Each patient had four MRI exam timepoints: pretreatment, early-treatment, mid-treatment, and post-treatment before and during their NAC.

In this study, we selected the T1 fat-saturated pre-contrast sequences from the original baseline MRI study. To mitigate the effects of surgical and pathological changes on breast density, we excluded any MRIs that were obtained following neoadjuvant treatment and evaluated only the contralateral normal breast for breast density analysis. Following preprocessing, **717** MRI cases remained for our analysis.

**Dataset 2: DBC-MRI:** This dataset includes 922 patients with biopsy-confirmed invasive breast cancer, collected from 2000 to 2014 in a single-institutional, retrospective study [41]. The dataset comprises extensive demographic, clinical, pathology, treatment, outcomes, and genomic data, sourced from clinical notes, radiology reports, and pathology reports. It has been instrumental in numerous published studies on radiogenomics and outcomes prediction. Consistent with the preprocessing approach used for Dataset 1 we excluded any MRIs obtained after treatment and focused on evaluating only the contralateral normal breast. This process resulted in a final set of **890** MRI cases for analysis.

**Dataset 3: Internal dataset at our institution:** The internal dataset consists of breast MRI and corresponding mammogram exams, collected from a single institution between March 2014 and 2021. This dataset includes breast MRI exams, along with their paired screening or diagnostic mammograms. For the purposes of this study, we selected a subset of exams based on specific criteria to focus on normal breast MRI cases with paired mammograms.

Initially, all available breast MRI exams conducted at our institution before December 31, 2021, were gathered. From this collection, a subset of **392** patients were identified using the following criteria: (1) screening examinations without a known cancer diagnosis, (2) no abnormal findings reported in the MRI, (3) had T1 fat-saturated pre-contrast sequences, and (4) mammograms taken within six months of the MRI. Criteria (1) and (2) were applied to minimize the influence of pathological changes on breast tissue, while criterion (3) was used to exclude redundant imaging data. Criterion (4) ensured that the mammogram and MRI were closely timed, minimizing the possibility of significant changes in breast density between the two exams. This selection process helped refine the dataset to include only normal, high-quality, and temporally aligned MRI-mammogram pairs, enabling a more accurate comparison between MR breast density calculations and radiologists' mammogram evaluations.

Notably, Dataset 3 comprises patients under screening, setting it apart from the previous two datasets where all patients had a known diagnosis of breast cancer. We aimed to determine whether the screened population exhibited a different breast density distribution compared to the other two groups.

4.2 Segmentation algorithm

In this study, we used the breast and dense tissue segmentation algorithm following the MR segmentation process we have previously reported [16]. This algorithm comprises two primary steps: the breast volume segmentation network and the breast dense tissue segmentation network. As shown in Figure 5, both the segmentation networks process single-channel MRI images to generate a breast and dense tissue volume mask. We employed a 3D V-Net [42] architecture for both segmentation tasks due to its residual connections and volumetric convolutions, which are particularly effective for modeling 3D anatomical structures. We also compared the performance of 3D V-Net with 3D U-Net and nnU-Net on the tissue segmentation task with DSC and Hausdorff Distance are employed as evaluation metrics. The detailed comparison is summarized in Table 2. Each network uses patch-based segmentation to optimize memory usage. The algorithm sequentially predicts small sub-volumes, which are then concatenated and fused to form the final segmentation output.

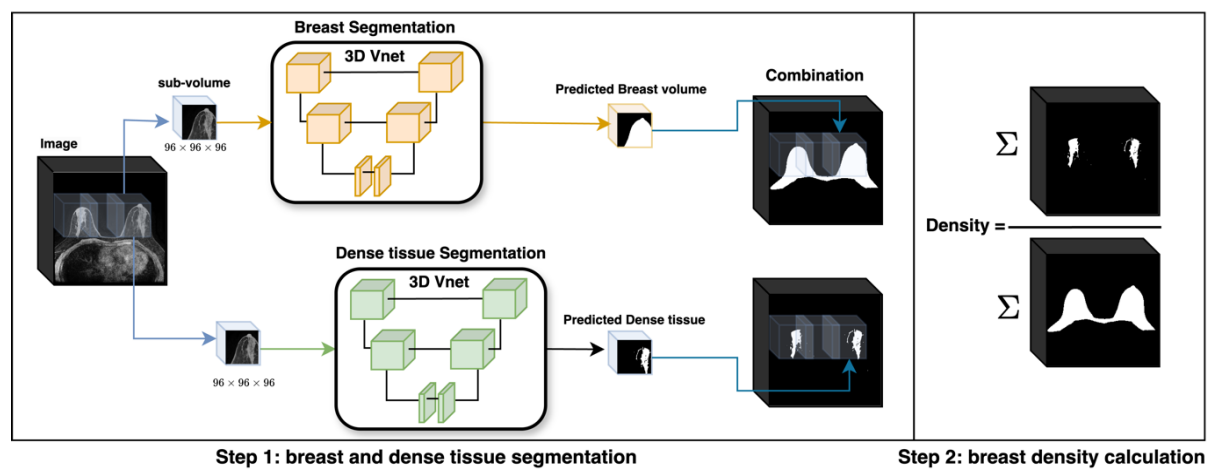

Figure 5: pipeline for breast density ratio calculation. It mainly contains two steps (1) Breast and dense tissue segmentation; (2) breast density ratio calculation based on two segmented masks.

In this work, we applied a 96x96x96 sub-volume for each inference, consistent with the training setting. The volume was divided into 8 uniform steps along the x and y directions and 3 uniform steps along the z direction during prediction. Predictions in overlapping regions were fused using an equal weighting (simple averaging) strategy, where each voxel's prediction is averaged across overlapping patches. Additionally, because these three datasets originate from different sources, we performed additional preprocessing steps based on their DICOM headers. Each dataset has a slightly different DICOM header format, so we adjusted our preprocessing accordingly to ensure optimal loading of the DICOM images as 3D volume in left-posterior-superior coordinate system. This also included reversing the order of the images in the Z direction for those exams with negative z position. After loading the dicom files as 3D volume, we also implemented a preprocessing as z-score normalization step before feeding the image to networks. Standardizing spatial orientation enhances feature consistency across different datasets, allowing the model to extract image patterns in a uniform manner. Meanwhile, z-score normalization mitigates scanner-induced intensity variations, improving cross-dataset consistency. The model evaluation is performed on an NVIDIA A6000 GPU, the dense tissue segmentation model takes approximately 5.07 seconds to process a single sample under the current settings.

### 4.3 Breast density calculation

After obtaining the segmented masks of breast volume (M_breast) and dense tissue (M_dense) for an MRI volume, we calculate this subject's breast density as the ratio of the area belongs to the dense tissue to the area of the whole breast:

$$\text{Breast density} = \frac{\sum M_{dense}}{\sum M_{breast}} \quad (1)$$

### 4.4 Comparing breast density in MRI with breast density in mammograms

To compare breast density calculated from MRI with the breast density provided by radiologists based on interpretation of DBT mammograms (hereafter referred to as mammograms for convenience) and their synthetic 2D images, we identified a cohort of patients which contained both mammograms and MRIs. The process to collect the paired mammogram and MRI is detailly described in Section 4.1 for Dataset 3: Internal dataset at our institution.

To determine the radiologist-assessed breast density from mammograms, we manually extracted key information from mammography reports and assigned each of the reports to one of the four categories: "Extremely dense","Heterogeneously dense", "Scattered areas of fibroglandular density", or "Almost entirely fatty". Mammogram reports containing terms such as "extremely dense" or "extremely fibroglandular" were classified as "Extremely dense." Reports mentioning "heterogeneously dense" or "heterogeneously fibroglandular" were grouped into the "Heterogeneously Dense" category. Those containing "scattered fibroglandular" were categorized under "Scattered areas of fibroglandular density," and mammogram results with terms like "entirely fatty" or "predominantly fatty" were assigned to the "Almost entirely fatty" category.

## 5 Acknowledgement

Research reported in this publication was supported by the National Institute Of Biomedical Imaging And Bioengineering of the National Institutes of Health under Award Number R01EB031575. The content is solely the responsibility of the authors and does not necessarily represent the official views of the National Institutes of Health.

We are grateful to the Duke Department of Surgery for providing support for this SPARK collaboration.

## 6 Data Availability Statement

Two out of the three datasets analyzed in this study are publicly accessible [39, 41]. However, the internal dataset is currently unavailable, as de-identifying the data requires an extensive institutional review process. Readers interested in evaluating the accuracy of the method can access and utilize the two publicly available datasets, which are readily accessible and user-friendly.

# 7 Code Availability Statement

We will release our code upon paper acceptance.

# 8 Completing Interests

Dr. Maciej A. Mazurowski is an Associate Professor at Duke University with appointments in Radiology, Electrical and Computer Engineering, Biostatistics and Bioinformatics, and Computer Science. He holds a 2015 patent (US-10980519-B2) for methods and systems designed for breast cancer detection. These methods and systems facilitate the determination of cancer lesion severity and disease prognosis by extracting and analyzing imaging features correlated with breast cancer subtypes and prognostic outcomes. All other authors do not have any financial or non-financial interests.

# 9 Author Contribution

Conceptualization, S.H., M.M., L.L., D.N., A.K.; data downloading, H.G.; data annotation, D.N,; model implementation, H.D.; data curation, Y.C., L.L.; methodology, Y.C., L.L., H.G.; formal analysis, Y.C., L.L., H.G.; paper writing, Y.C., H.G., S.H., M.M., L.L., D.N; all authors have reviewed and agreed to the published version of the manuscript. Y.C. and L.L. contributed equally.